# PROBING THE HIDDEN TALENT OF ASR FOUNDATION MODELS FOR L2 ENGLISH ORAL ASSESSMENT

*Fu-An Chao, Bi-Cheng Yan, Berlin Chen*

National Taiwan Normal University, Taipei, Taiwan

{fuanchao, bicheng, berlin}@ntnu.edu.tw

## ABSTRACT

In this paper, we explore the untapped potential of Whisper [1], a well-established automatic speech recognition (ASR) foundation model, in the context of L2 spoken language assessment (SLA). Unlike prior studies that *extrinsically* analyze transcriptions produced by Whisper, our approach goes a step further to probe its *latent* capabilities by extracting acoustic and linguistic features from hidden representations. With only a lightweight classifier being trained on top of Whisper's intermediate and final outputs, our method achieves strong performance on the GEPT picture-description dataset, outperforming existing cutting-edge baselines, including a multimodal approach. Furthermore, by incorporating image and text-prompt information as auxiliary relevance cues, we demonstrate additional performance gains. Finally, we conduct an in-depth analysis of Whisper's embeddings, which reveals that, even without task-specific fine-tuning, the model intrinsically encodes both ordinal proficiency patterns and semantic aspects of speech, highlighting its potential as a powerful foundation for SLA and other spoken language understanding tasks.

## 1. INTRODUCTION

In recent years, there has been a surge of interest in the emergent abilities [2] of large-scale pre-trained foundation models. These abilities, which were not explicitly targeted during training, arise once the model reaches a sufficient scale of data and parameters. A hallmark of such models is their capacity for zero-shot transfer [3], allowing them to tackle previously unseen tasks without task-specific fine-tuning. Fundamentally, such versatility has reshaped how researchers approach a wide range of complex problems across both the NLP [4] and vision [5] communities, while also catalyzing new research directions such as in-context learning and chain-of-thought prompting to better steer model behavior [6].

While significant advances have been made in text and vision models, the exploration of zero-shot capabilities in speech-based foundation models remains relatively limited. Most existing efforts have centered on Whisper; a weakly supervised speech recognition model trained on 680k hours of multilingual and multitask data. For instance, Peng et al. [7] leveraged prompt engineering to adapt Whisper for zero-shot task generalization, while Li et al. [8] introduced open-vocabulary keyword-spotting combining crafted prompts for contextual biasing. Although both approaches yielded substantial improvements, the scope of their findings was restricted to tasks already near Whisper's pre-training domains, specifically speech recognition and speech translation. On the other hand, [9] investigated template-based text prompts and task calibration on 8 audio-classification datasets, showing that debiasing can unlock Whisper's zero-shot classification potential. In contrast to the above studies, Whisper-AT [10] explored the encoder rather than the decoder and found that its audio representations are noise-aware. Building on this, the authors trained a unified ASR and audio-tagging model that delivered strong performance. However, few studies have considered examining both the encoder and decoder in tandem, leaving open questions about their underlying synergies.

In addition, due to its pre-training, Whisper's input length is capped at 30 seconds. This limitation, present even in its larger variants, not only constrains the research tasks that can be explored but also leaves much of the model's hidden potential untapped. For those tasks exceeding this window, such as long-context understanding, processing must rely on decoding the transcribed text through sequential [1] or chunked algorithms [11]. In such cases, only the final textual output is accessible, while the rich acoustic information within the model remains largely out of reach.

To bridge these gaps, we tap into Whisper for use in spoken language assessment (SLA), a challenging task that requires long-context understanding. Instead of depending solely on final transcriptions, we delve into the model's hidden representations, extracting rich acoustic and linguistic embeddings from both the Whisper encoder and decoder through a simple chunking and hierarchical pooling strategy for full-context modeling. On top of these features, we train only a lightweight classifier, which nonetheless proves sufficient for effective prediction. Furthermore, we demonstrate that the performance can be further improved by injecting the image and text-prompt information to better capture content relevance, underscoring the flexibility of our approach in integrating auxiliary signals. Experiments on the GEPT picture-description dataset show promising classification results, surpassing existing advanced approaches and revealing Whisper's unexplored potential for SLA and broader spoken language understanding tasks.

## 2. PROPOSED METHOD

Existing work applying Whisper to SLA has primarily revolved around its ASR capabilities, followed by either error analysis [12] or modeling [13] of the decoded text. In contrast, our approach treats Whisper as a frozen feature extractor, leveraging its hidden representations to obtain acoustic and linguistic features for downstream holistic score prediction. As illustrated in Figure 1, our framework consists of two stages: (a) feature extraction and (b) classifier training.

### 2.1. Feature extraction

As a hard 30-second input limit is baked into Whisper, long-form audio is by default truncated to the first 30 seconds. To overcome this constraint, we develop a simple chunking algorithm, similar to [11], that makes full use of the entire audio signal. We refer to this pre-processing step in feature extraction as *segmentation*.

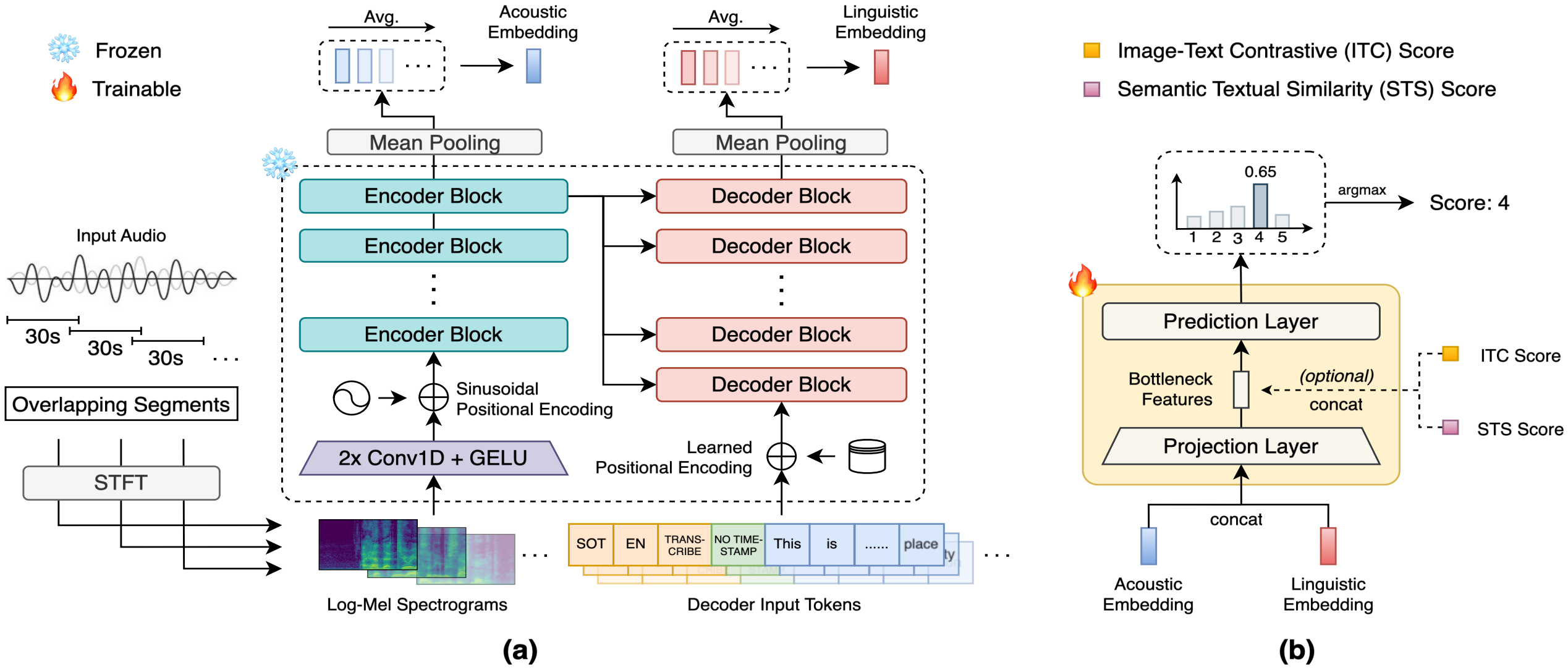

**Fig. 1.** Overview of the proposed approach, comprising two stages: (a) feature extraction and (b) classifier training.

**Segmentation.** Given a sequential input signal $\mathbf{x} \in \mathbb{R}^N$ where $N$ denotes the number of samples, we first split $\mathbf{x}$ into fixed-length segments of size $L$ with stride $S$, where $S < L$ ensures an overlap of $O = L - S$. This procedure yields $K$ equal-sized chunks $\mathbf{c}_i \in \mathbb{R}^L$, $i = 1, \ldots, K$, where $K = \lfloor \frac{N-L}{S} \rfloor + 1$, such that only complete chunks of length $L$ are retained. Each chunk $\mathbf{c}_i$ is then transformed into a log-Mel spectrogram $\mathbf{X}_i \in \mathbb{R}^{F \times M}$, where $F$ is the number of time frames and $M$ the number of mel bins, via the short-time Fourier transform (STFT), and subsequently fed into Whisper for feature extraction.

**Acoustic features.** Being known to capture rich acoustic information [10], we extract acoustic features from the Whisper encoder. For each input $\mathbf{X}_i$, we compute a chunk-level acoustic embedding $\bar{\mathbf{h}}_i^{\text{enc}}$ as follows:

$$\mathbf{H}_i^{\text{enc}_0} = \text{ConvolutionLayer}(\mathbf{X}_i) + \mathbf{P}^{\text{enc}}, \tag{1}$$

$$\mathbf{H}_i^{\text{enc}_\text{N}} = \text{Encoder}(\mathbf{H}_i^{\text{enc}_0}), \tag{2}$$

$$\bar{\mathbf{h}}_i^{\text{enc}} = \text{MeanPooling}(\mathbf{H}_i^{\text{enc}_\text{N}}), \tag{3}$$

where $\text{ConvolutionLayer}(\cdot)$ comprises two 1-D convolutions with GELU activations, one employing a stride of 2, while $\mathbf{P}^{\text{enc}}$ denotes the sinusoidal position embeddings. Each chunk embedding $\bar{\mathbf{h}}_i^{\text{enc}} \in \mathbb{R}^d$ is obtained from the encoder's last hidden states $\mathbf{H}_i^{\text{enc}_\text{N}} \in \mathbb{R}^{F/2 \times d}$ by applying mean pooling across the time frames. Finally, we aggregate the chunk-level embeddings into a global utterance-level acoustic representation:

$$\mathbf{v}^{\text{enc}} = \text{MeanPooling}(\{\bar{\mathbf{h}}_i^{\text{enc}}\}_{i=1}^K). \tag{4}$$

Given that the first pooling step compresses each chunk into a fixed-length vector, and the second aggregates across chunks to produce a single utterance-level vector, this two-stage process is referred to as a *hierarchical pooling* strategy.

**Linguistic features.** Since the Whisper decoder is trained as an autoregressive conditional language model, it requires the decoder input tokens for feature extraction. Each decoding sequence begins with a fixed prefix (e.g., <|startoftranscript|> token) and is extended with tokens generated autoregressively. However, autoregressive generation is computationally expensive when the aim is merely to extract features. To bypass this, for each chunk $\mathbf{c}_i$, we construct the decoder input tokens $\mathbf{z}_i$ by concatenating its transcription tokens $\boldsymbol{\tau}_i$ with the required prefix tokens $\mathbf{p}_i$:

$$\mathbf{z}_i = [\mathbf{p}_i; \boldsymbol{\tau}_i]). \tag{5}$$

This provides a complete decoder input without the overhead of autoregressive decoding, enabling efficient extraction of decoder-side linguistic embeddings. A key advantage of this approach is its flexibility: the transcription tokens $\boldsymbol{\tau}_i$ can be obtained from any ASR backbone, allowing the extracted linguistic embeddings to benefit from strong recognition models when ground-truth transcripts are unavailable. Drawing an analogy to teacher forcing [14] but adapting it for inference, we term this approach *pseudo-teacher forcing*. Given $\mathbf{z}_i$, a chunk-level linguistic embedding $\bar{\mathbf{h}}_i^{\text{dec}}$ is then extracted as follows:

$$\mathbf{H}_i^{\text{dec}_0} = \text{Embedding}(\mathbf{z}_i) + \mathbf{P}^{\text{dec}}, \tag{6}$$

$$\mathbf{H}_i^{\text{dec}_M} = \text{Decoder}\left(\mathbf{H}_i^{\text{dec}_0}, \mathbf{H}_i^{\text{enc}}\right), \tag{7}$$

$$\bar{\mathbf{h}}_i^{\text{dec}} = \text{MeanPooling}\left(\mathbf{H}_i^{\text{dec}_\text{M}}\right), \tag{8}$$

where $\text{Embedding}(\cdot)$ is the token embedding layer, $\mathbf{P}^{\text{dec}}$ denotes the learned position embeddings. $\mathbf{H}_i^{\text{dec}_M}$ are the last hidden states of the decoder. Finally, the utterance-level linguistic representation is obtained by aggregating across all chunks:

$$\mathbf{v}^{\text{dec}} = \text{MeanPooling}(\{\bar{\mathbf{h}}_i^{\text{dec}}\}_{i=1}^K). \tag{9}$$

**Auxiliary features.** To better assess L2 learners' language competence in different facets, SLA tasks are often designed as multi-level monologues (e.g., reading aloud or picture description). In such contexts, auxiliary information like text prompts and images can provide valuable cues for evaluation beyond what is captured in the speech signal alone. To this end, we incorporate two extra features: STS and ITC scores, for measuring prompt coherence and image relevance, respectively, as depicted in Figure 2.

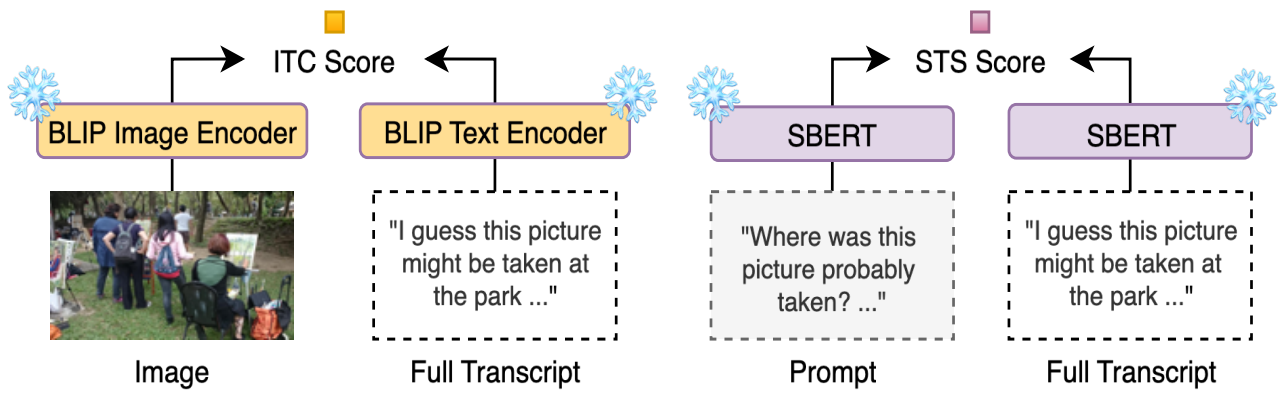

**Fig. 2.** An illustration of the auxiliary feature extraction.

**Semantic textual similarity (STS) score:** We compute STS [15] to quantify the semantic coherence between the given text prompt $Q$ and the learner's response $T$:

$$\mathbf{e}_Q = \text{SBERT}(Q), \quad \mathbf{e}_T = \text{SBERT}(T), \tag{10}$$

$$\mathbf{s}_{\text{STS}} = \mathbf{e}_Q \cdot \mathbf{e}_T, \tag{11}$$

where $\mathbf{e}_Q$ and $\mathbf{e}_T$ are embeddings generated using a pre-trained SBERT model [16]. Since the specific model[1] we adopt is trained with dot-product similarity, we directly use the dot score as the STS measure.

**Image-text contrastive (ITC) score:** To evaluate the relevance between learner responses and visual prompts, we employ BLIP[2] [17], a vision-language foundation model jointly pre-trained with three objectives: image-text contrastive (ITC), image-text matching (ITM), and language modeling (LM). Here, we adopt the ITC objective to measure image relevance, where the score is defined as the cosine similarity between the BLIP-encoded embeddings of the image $I$ and learner's response $T$:

$$\mathbf{b}_{img} = \text{BLIP}_{img}(I), \quad \mathbf{b}_{txt} = \text{BLIP}_{txt}(T), \tag{12}$$

$$\mathbf{s}_{\text{ITC}} = \cos(\mathbf{b}_{img}, \mathbf{b}_{txt}) = \frac{\mathbf{b}_{img} \cdot \mathbf{b}_{txt}}{\|\mathbf{b}_{img}\| \|\mathbf{b}_{txt}\|}. \tag{13}$$

### 2.2. Classifier training

During the classifier training, the acoustic embedding $\mathbf{v}^{\text{enc}}$ and linguistic embedding $\mathbf{v}^{\text{dec}}$ are concatenated and projected into a compact bottleneck feature space [18]:

$$\mathbf{v}^{\text{bnf}} = f_{\text{proj}}([\mathbf{v}^{\text{enc}}; \mathbf{v}^{\text{dec}}]). \tag{14}$$

To enrich $\mathbf{v}^{\text{bnf}}$, the STS score $\mathbf{s}_{\text{STS}}$ and the ITC score $\mathbf{s}_{\text{ITC}}$ are optionally appended, forming a fuse representation:

$$\mathbf{u} = [\mathbf{v}^{\text{bnf}}; \mathbf{s}_{\text{STS}}; \mathbf{s}_{\text{ITC}}]. \tag{15}$$

The prediction layer then produces logits:

$$\mathbf{o} = f_{\text{pred}}(\mathbf{u}). \tag{16}$$

from which the proficiency probabilities are obtained via. $\hat{\mathbf{y}} = \text{softmax}(\mathbf{o})$, and the model parameters are optimized using cross-entropy loss between $\hat{\mathbf{y}}$ and the ground-truth label $\mathbf{y}$. Overall, this architecture integrates both primary embeddings and auxiliary scores, guiding the classifier to capture dimensions aligned with key aspects of standardized scoring rubrics [19], including delivery quality, linguistic accuracy, and content relevance.

[1] https://huggingface.co/sentence-transformers/multi-qa-mpnet-base-dot-v1
[2] https://huggingface.co/Salesforce/blip-itm-large-flickr

**Table 1**. Performance impact of the segmentation on Whisper.

| Methods | Seen test | | Unseen test | |
|---|---|---|---|---|
| | Weighted-F1 | Acc. | Weighted-F1 | Acc. |
| WhisperEncoder [1] | 0.648 | 0.678 | 0.689 | 0.710 |
| **Ours (acoustic)** | **0.683** | **0.722** | **0.709** | **0.723** |

**Table 2**. Performance comparison of different features. (PTF: pseudo-teacher forcing, ALL: acoustic+linguistic+auxiliary)

| Methods | Seen test | | Unseen test | |
|---|---|---|---|---|
| | Weighted-F1 | Acc. | Weighted-F1 | Acc. |
| wav2vec 2.0 [22] | 0.557 | 0.567 | 0.602 | 0.617 |
| **Ours (acoustic)** | **0.683** | **0.722** | **0.709** | **0.723** |
| BERT [21] | 0.559 | 0.578 | 0.659 | 0.680 |
| **Ours (linguistic)** | **0.660** | **0.678** | **0.726** | **0.740** |
| w/o PTF | 0.633 | 0.655 | 0.715 | 0.720 |
| **Ours (acou.+ling.)** | **0.709** | **0.733** | **0.751** | **0.757** |
| **Ours (ALL)** | **0.742** | **0.767** | **0.762** | **0.760** |
| w/o ITC Score | 0.720 | 0.744 | 0.756 | 0.759 |
| w/o STS Score | 0.729 | 0.744 | 0.715 | 0.710 |

## 3. EXPERIMENTS

### 3.1. Experimental setup

We evaluated our approach on the GEPT picture-description dataset [13], which contains authentic spoken responses (≈85s each) to image-based prompts for intermediate-level English assessment. Following [13], fractional holistic scores are rounded down to a discrete 1-5 scale for training, where scores>3 indicate performance above the CEFR B1 level, whereas scores≤3 denote failure. The dataset is split into: train (N=719), dev (N=90), seen test (N=90), and unseen test (N=300) sets, supporting evaluation on both seen and unseen prompts. Notably, part of the dataset provides sub-score annotations; we use relevance scores for analysis and report weighted F1, accuracy, and binary accuracy as evaluation metrics.

According to [20], we adopt Whisper-medium as the backbone and segment audio into 30-s segments with 5-s overlap for feature extraction. To facilitate inference, we use Distil-Whisper[3] [11] as the teacher model for pseudo-teacher forcing. The projection layer $f_{\text{proj}}$ has a hidden size of 512, and the classifier is trained for 1k steps with a learning rate of 7.5e-4, batch size 4 and gradient accumulation of 2. To ensure determinism, all experiments are conducted with a fixed random seed. The source code will be made publicly available in the camera-ready version.

### 3.2. Compared methods

In this work, we compare our approach with several strong baselines, grouped into two categories: single-modal and multi-modal methods. The single-modal baselines include a text-only model based on BERT [21] and a speech-only model based on wav2vec 2.0 [22], both kept frozen during training. For multi-modal baselines, we consider the joint use of BERT and wav2vec 2.0 [22], SAMAD [13], and a recent multifaceted approach [24].

[3] https://huggingface.co/distil-whisper/distil-large-v3.5

**Table 3**. Performance evaluations of our model and several multi-modal baselines. (A: audio, V: vision, T: text;  N/A: not available)

| Methods | Year | Modality | Seen test | | | Unseen test | | |
|---|---|---|---|---|---|---|---|---|
| | | | Weighted-F1 | Acc. | Bin. Acc. | Weighted-F1 | Acc. | Bin. Acc. |
| wav2vec2.0+BERT [23] | 2023 | A+T | 0.639 | 0.644 | N/A | 0.650 | 0.667 | N/A |
| SAMAD [13] | 2024 | A+T | 0.648 | 0.656 | N/A | 0.684 | 0.697 | N/A |
| Lu et al. [24] | 2025 | A+V+T | N/A | 0.700 | 0.789 | N/A | 0.717 | 0.797 |
| **Ours** | 2025 | A+V+T | **0.742** | **0.767** | **0.889** | **0.762** | **0.760** | **0.837** |

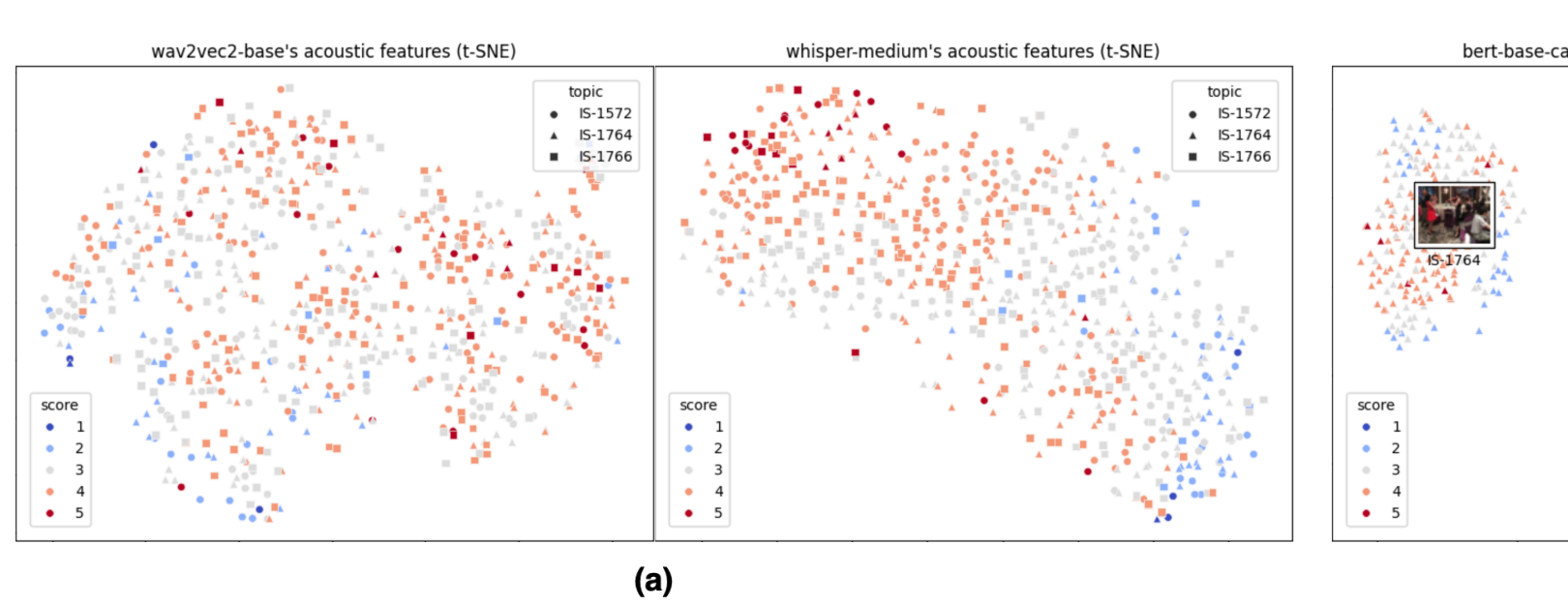

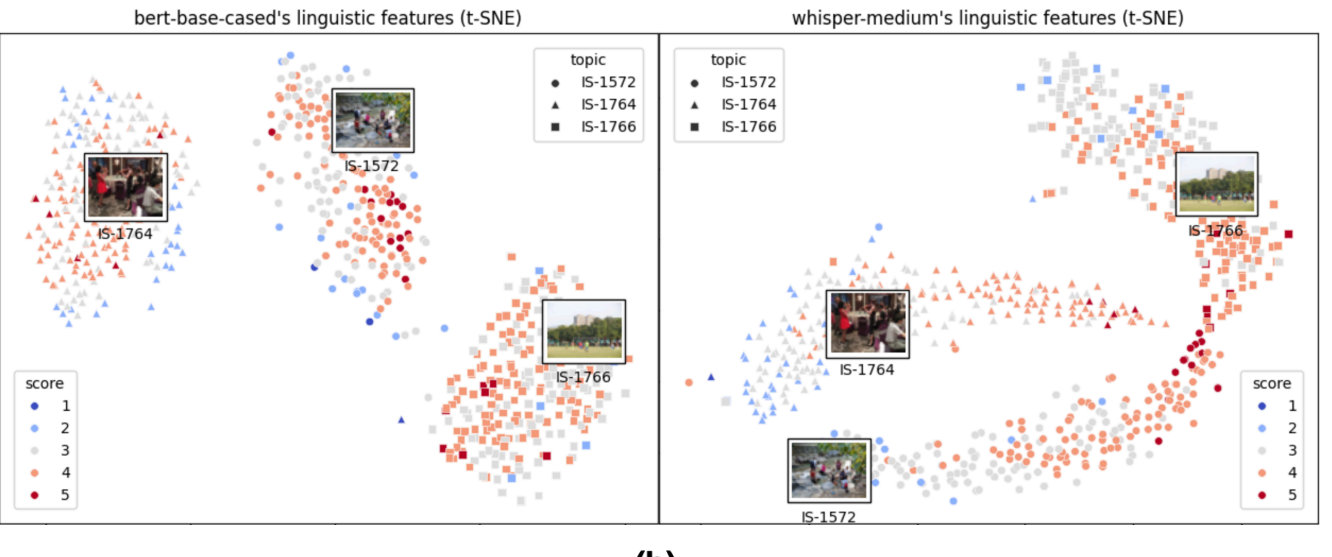

**Fig. 3.** t-SNE visualizations of (a) acoustic and (b) linguistic embeddings, where each point represents an utterance-level representation.

### 3.3. Performance comparison

**Segment or not?** In our preliminary experiments, we focused on the whisper encoder to investigate the impact of the chunking strategy on performance. Specifically, we compared the classification results using $\bar{\mathbf{h}}_1^{\text{enc}}$ and $\mathbf{v}^{\text{enc}}$. As shown in Table 1, WhisperEncoder denotes the original settings, which processes only the first 30 seconds of an utterance in a single pass. Segmenting the audio into overlapping 30-second chunks allows Whisper to exploit the entire context, leading to higher classification accuracy.

**Feature ablations.** To better understand the Whisper's latent capacities and the proposed auxiliary features, we conducted a series of feature ablation experiments, as summarized in Table 2.

**Acoustic features:** Our proposed acoustic features $\mathbf{v}^{\text{enc}}$ surpass wav2vec 2.0. As shown in Table 1 and Table 2, Whisper consistently outperforms wav2vec 2.0, even with only the first 30 seconds of audio, suggesting that its large-scale multilingual pre-training enables it to extract salient acoustic cues from shorter audio clips and generalize more effectively than wav2vec 2.0.

**Linguistic features:** Similarly, our proposed linguistic features $\mathbf{v}^{\text{dec}}$ outperform BERT, as demonstrated in Table 2, indicating the rich linguistic information encoded in Whisper's decoder. We further enhance $\mathbf{v}^{\text{dec}}$ by employing proposed pseudo-teacher forcing strategy, which leverages knowledge distilled from a large ASR model to boost performance. Finally, integrating $\mathbf{v}^{\text{enc}}$ and $\mathbf{v}^{\text{dec}}$ to form $\mathbf{v}^{\text{bnf}}$ (*c.f.* Eq. (14)) yields even better results. This is because the two features are complementary: $\mathbf{v}^{\text{enc}}$ excels with seen prompts, while $\mathbf{v}^{\text{dec}}$ is superior for unseen prompts.

**Auxiliary features:** We introduce auxiliary features into $\mathbf{v}^{\text{bnf}}$ to create $\mathbf{u}$ (*c.f.* Eq. (15)), which yields additional performance gains for classification, particularly on seen prompts (see Table 2). By dropping each feature individually, we determined that ITC mainly contributes to overall robustness, while STS is particularly crucial for maintaining high performance on unseen prompts.

**Overall performance.** In comparison to existing state-of-the-art multimodal baselines (see Table 3), the proposed method obtains significant improvements in classification performance. The binary accuracy (pass/fail) on both test sets further confirms that our approach is more robust for standardized testing scenarios.

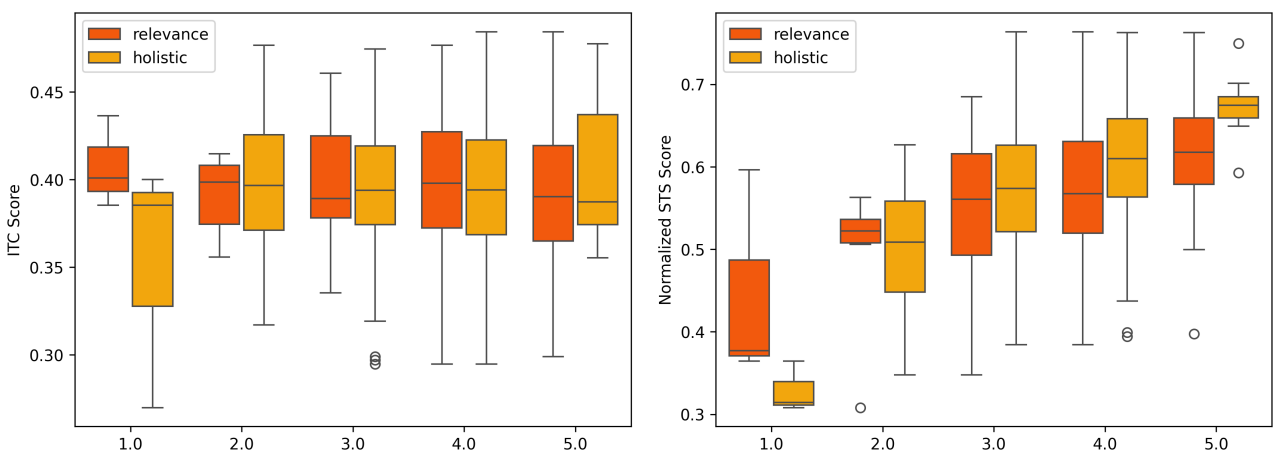

**Fig. 4.** ITC and STS scores against relevance and holistic scores.

### 3.4. Feature visualization analysis

To probe the source of system performance, we visualized the proposed features in Figures 3 and 4. As shown in Figure 3, Whisper's acoustic embeddings exhibit a more pronounced ordinal alignment with proficiency scores, whereas wav2vec 2.0 produces more ambiguous, diffuse clusters that are less sensitive to score variation. This may suggest that Whisper's multitask training better preserves the fluency [25] and prosodic [26] cues tied to assessment. For linguistic embeddings, BERT shows strong semantic clustering by topic but lacks ordinal separation, reflecting its text-centric pre-training. In contrast, Whisper's linguistic embeddings not only retain topic-based structure but also reveal score-related gradients, likely due to its audio-conditioned nature (*c.f.* Eq. (7)), thereby inheriting ordinality from acoustic features.

Figure 4 further illustrates the roles of the auxiliary features. For holistic scores, STS serves as a better measure of overall response quality, while ITC is a strong indicator in identifying off-topic or low-quality samples (score = 1). In comparison, relevance scores from human raters emphasize prompt coherence (STS) rather than strict image relevance (ITC).

## 4. CONCLUSION

In this paper, we have demonstrated that Whisper, beyond its role as an ASR system, provides rich acoustic and linguistic representations that can be leveraged for SLA. This study marks an initial step, and we envisage future work that not only extends to multimodal settings but also explores generating rationales for scores, thereby moving this line of research toward explainable AI in speaking assessment.

## 5. ACKNOWLEDGMENT

This work was supported by the Language Training and Testing Center (LTTC), Taiwan. Any findings and implications in the paper do not necessarily reflect those of the sponsor.